\documentclass[journal]{IEEEtran}

\usepackage{times,amsmath,float,color,subfigure,graphicx,epstopdf,textcomp,gensymb}
\usepackage[table,xcdraw]{xcolor}
\usepackage[noadjust]{cite}

\usepackage{epsfig,textcomp,amsfonts,multirow}
\usepackage{array,color,MnSymbol}
\usepackage{etoolbox,siunitx}
\usepackage{bigstrut}

\usepackage{cite} 

\usepackage{hyperref}
\hypersetup{
  unicode=false,     
  pdftoolbar=true,    
  pdfmenubar=true,    
  pdffitwindow=false,   
  pdfstartview={FitH},  
  pdfauthor={},   
  pdfcreator={},  
  pdfproducer={}, 
  pdfkeywords={}{}, 
  pdfnewwindow=true,   
  colorlinks=true,    
  linkcolor={blue},     
  citecolor=blue,     
  filecolor=blue,     
  urlcolor=blue      
}
\newcommand{\ba}{{\boldsymbol \alpha}}
\newcommand{\myphi}{{\boldsymbol \varphi}}

\begin{document}
\title{Graph Embedding via High Dimensional Model Representation for Hyperspectral Images}

\author{G\"ul\c sen Ta\c sk\i n,~\IEEEmembership{Member,~IEEE}
        and~Gustau Camps-Valls,~\IEEEmembership{Fellow,~IEEE}
\thanks{{GT is with the Institute of Disaster Management, Istanbul Technical University, Turkey. GCV is with the Image Processing Laboratory, Universitat de Val\`encia, Spain.}}
\thanks{{This work was supported by the Scientific and Technological Research Council of Turkey (T\"UBITAK)-1001 (Project number: 217E032). Gustau Camps-Valls was supported by the European Research Council (ERC) under the ERC Consolidator grant project SEDAL (grant agreement 647423).}}
\thanks{\color{black}\copyright 2021 IEEE.  Personal use of this material is permitted.  Permission from IEEE must be obtained for all other uses, in any current or future media, including reprinting/republishing this material for advertising or promotional purposes, creating new collective works, for resale or redistribution to servers or lists, or reuse of any copyrighted component of this work in other works.}}



\markboth{ IEEE Transactions on Geoscience and Remote Sensing}
{Gulsen Taskin}

\maketitle
\begin{abstract}
\textcolor{black}{Learning the manifold structure of remote sensing images is of paramount relevance for modeling and understanding processes, as well as to encapsulate the high dimensionality in a reduced set of informative features for subsequent classification, regression, or unmixing.} Manifold learning methods have shown excellent performance to deal with hyperspectral image (HSI) analysis but, unless specifically designed, they cannot provide an explicit embedding map readily applicable to out-of-sample data. A common assumption to deal with the problem is that the transformation between the \textcolor{black}{high-dimensional input space} and the (typically low) latent space is linear. This is a particularly strong assumption, especially when dealing with hyperspectral images due to the well-known nonlinear nature of the data. To address this problem, a manifold learning method based on High Dimensional Model Representation (HDMR) is proposed, which enables to present a nonlinear embedding function to project out-of-sample samples into the latent space. The proposed method is compared to manifold learning methods along with its linear counterparts and achieves promising performance in terms of classification accuracy of \textcolor{black}{a representative set of} hyperspectral images.
\end{abstract}


\section{Introduction}
\label{sec:intro}
Hyperspectral images (HSI) provide very rich spectral information, and help many purposes, from image classification to anomaly detection and parameter retrieval. Information extraction from hyperspectral images, however, is a challenging task \textcolor{black}{in all different steps along} the standard remote sensing image processing chain, such as data processing, storage, and transmission  \cite{Pasolli2016,CampsValls201445}. The high dimensionality of HSI poses the well-known problem of the curse of dimensionality, \textcolor{black}{also} known as the Hughes phenomenon, which leads to problems of collinearity and challenges the use of automatic machine learning models \cite{Kucuk2016,Taskin2019a}. 
To cope with all this, it is customary and very effective to reduce the dimensionality of hyperspectral data with all kinds of projection-on-subspaces methods, which try to summarize the data into a reduced set of components, while still \textcolor{black}{accounting for} the underlying structure of the original data in an efficient way. Dimensionality reduction (DR) methods constitute a useful approach to learning feature representations of the original data. Methods can be either linear or nonlinear and can try to find either explicit or implicit transformation. In general, linear DR methods, such as PCA \cite{Jolliffe:1986}, random projection \cite{Bingham}, and LDA \cite{Fukunaga1990}, are found to be computationally efficient, easy to use and apply, and generally find useful and effective data projections, but their ability in embedding is limited especially when processing high-dimensional nonlinear data \cite{Yin2016a, Bingham}. 
\textcolor{black}{Even though linearity is a convenient assumption, it is far from being realistic. Actually, 
nonlinearity} appears in different forms in remote-sensing data \cite{Bachmann2005,Bachmann2006}: 1) the nonlinear scatter described in the bidirectional reflectance distribution function; 2) the variable presence of water in pixels as a function of position in the landscape; 3) multiscattering and heterogeneities at the subpixel level; 4) atmospheric and geometric corrections that need to be done to deliver useful image products. These facts need to be encoded in the method either by allowing flexible nonlinear mappings or by an accurate description of the data-manifold coordinates \cite{functionalAnalysis1,functionalAnalysis2}.

Graph embedding (GE) is a very effective method for dimensionality reduction providing data-manifold coordinates of the samples lying in a high dimensional space. The GE represents each sample of training data as a vertex in an undirected graph, and the similarity of the connected vertices is determined based on a statistic or geometric properties of the corresponding samples, denoted as a graph affinity matrix. The aim of GE is to find an optimal mapping from a high dimensional to a low dimensional space by preserving the local or global structure of the manifold, that is captured by the affinity matrix. Each vertex is projected into low-dimensional space with eigenvectors of corresponding eigenvalues of the Laplacian matrix \cite{Yin2016a}. Yan et al. \cite{Yan2007} showed that many of the common methods, including 
LDA \cite{Fukunaga1990}, Laplacian Eigenmaps \cite{Niyogi2003},  Locally Linear Embedding (LLE) \cite{Roweis2000}, Locality Preserving Projections (LPP) \cite{He2004}, Neighborhood Preserving Embedding (NPE) \cite{He2005a}, Kernel Discriminant Analysis (KDA) \cite{Baudat2000}, Kernel LPP, \cite{He2004} and Kernel Eigenmaps \cite{Brand2003aug}, can be expressed in the form of graph embedding (GE) along with their {\em linearization, kernelization and tensorization.} Therefore, the graph embedding framework is highly flexible in terms of developing new DR methods, that are ensured simply by designing a specific affinity matrix or modifying the optimization function by adding a regularizer or a constraint. 

As the affinity matrix plays an important role in the performance of the graph embedding, several attempts have been made to construct a more sophisticated affinity matrix to exploit the underlying manifold. Depending on the choice of the affinity matrix, the GE framework leads to many different types of dimensionality reduction methods. Hang and Liu \cite{Hang2018a} suggested a method of regularization to incorporate the spatial information by designing an intraclass graph for each nonoverlapping superpixel generated by the method of over-segmentation. Instead of using a fixed affinity matrix, \cite{Wang2018b}  proposed a method to generate an optimal graph by adjusting the neighborhood relationship. The optimal graph and ridge regressor was jointly conducted to learn an embedding function that can be extended to new data.  Camps-Valls et. al  proposed a graph kernel function for spatio-spectral remote sensing image classification \cite{CampsValls2010741}. To represent the complex multiple relationships of HSI, a hypergraph model was presented by \cite{Luo2018} to incorporate spatial information more efficiently, and thus an optimal projection matrix is obtained based on the feature learning model, that is designed to compact the intraclass properties and separate the interclass similarities of the HSI. While several other methods have been introduced based on a graph affinity matrix \cite{HuangShi2020}, a certain number of studies rely more on developing supervised or semi-supervised DR methods utilizing GE framework \cite{CampsValls20073044}. 

However, like in kernel machines, \textcolor{black}{computational} efficiency is compromised and adoption in real practice has been questioned because graph embedding methods do not provide a straightforward out-of-sample (OOS) prediction function in general unless specifically designed. Direct graph embedding provides a low-dimensional representation only for the training samples and can be extended for all the samples by using some linear or nonlinear approaches \cite{HongYokoya2020}. Among them, the work conducted by \cite{Bengio2003} is the very well-known nonlinear approach providing a general solution to the out-of-sample problem for a certain type of unsupervised ML methods based on the Nystr\"om formula. 

The other approaches rely on linearization or kernelization of the graph embedding framework, resulting in a specific manifold learning method, which is also capable of solving the out-of-sample problem. Among them, the Locality Preserving Projection (LPP) \cite{He2004} is one of the linear DR methods derived by modifying the objective function of Laplacian Eigenmaps to provide a projection matrix based on the assumption that transformation between the high and the low dimensional latent space is linear. Wang et al. \cite{Wang2017d} extended the LPP approach to minimize the locality and maximized the globality simultaneously under the orthogonal constraint.  Huang et al. \cite{Huang2015d} proposed a sparse manifold embedding method based on the graph embedding and sparse representation, which overcomes the out-of-sample problem. Neighborhood Preserving Embedding (NPE) is another method using the linearity assumption but an extension for LLE \cite{He2005a}. However, such linear assumptions might be too restrictive, especially for datasets having inherent nonlinearities. 

The direct use in hyperspectral image processing, for example, typically results in poor data representations for subspace learning \cite{DeMorsier2016}. Based on the motivation of incorporating nonlinearity into the dimensionality reduction, Hang and et. al proposed a semi-supervised graph regularized ridge regression by also considering its kernelized version to deal with nonlinear feature relations in remote sensing data \cite{Hang2017}. With a similar inspiration, Tanga et. al. introduced a graph regularized dimensionality reduction method for the  OOS problem since neither the low-rank matrix approximation with manifold regularization nor graph-Laplacian PCA is capable of handling new samples \cite{Tang2017}. Liu and Ma \cite{Liu2018} proposed a locality preserving method by assuming that the low-dimensional coordinates of the samples are reconstructed with Gaussian coefficients. Since almost all the methods described above deal with the OOS problem in an unsupervised setting, Vural and Guillemot introduced a generalized solution to the OOS problem for supervised manifold learning for classification,  involving a radial basis function interpolator by minimizing the total embedding error with a regularization term \cite{Vural2015}. Hong et al. \cite{Qiao2013} proposed an explicit nonlinear polynomial mapping based on the graph embedding from the high dimensional space to low-dimensional space. Sparks and Madabhushi developed a method for mapping the images never seen before to the low-dimensional space by combining out-of-sample extrapolation with semi-supervised learning in a unified framework \cite{Sparks2016}.  Dornika and El Traboulsi introduced a  flexible graph-based semi-supervised embedding as well as its kernelized version for classification tasks, which can be generalized to the entire high-dimensional input space \cite{Dornaika2016}.

The literature shows that there is a large variety of studies pointing out graph-based dimensionality reduction and its out-of-sample problem. Based on this motivation, in this paper, a manifold learning method utilizing High Dimensional Model Representation (HDMR) is proposed to extend the linearity assumption in graph embedding to the nonlinearity as a solution to out-of-sample problems as well. 
The HDMR is a mathematical model representing an $n-$dimensional multivariate function as a linear combination of $n$ low-order component functions \cite{Rabitz1999b}, \textcolor{black}{which has been} 
used in many different application areas \cite{Erten2019,Taskin2017,Tunga2018,Rahman2011b,Chen2019}. {\color{black}
This paper extends our work in \cite{tacskin2020manifold} by  1) introducing HDMR in the graph embedding framework for dimensionality reduction with a more theoretical analysis, 2) providing an explicit function to work with out-of-sample embeddings, and 3) illustrating its wide applicability in hyperspectral image feature extraction and classification. In our proposed method, we use Legendre Polynomials (LPs) in the approximation of the embedding function via HDMR, mainly because of their simple derivation from monomial basis via Gram-Schmidt orthogonalization \cite{boyd2014relationships}.  Moreover, the LPs are more suitable than orthogonalized B-splines and Chebyshev polynomials as they are more suitable with the unit weights we use in this study \cite{rahman2020spline}.}

We evaluate the performance of our method in three well-curated and standard hyperspectral datasets 
via classification on the projected subspace. The results show that the proposed method outperforms other alternatives, and even achieves higher classification accuracy when using all the available features in the classification.

\section{High Dimensional Model Representation}

The HDMR expansion of a multivariate function $f$ \textcolor{black}{working on data points ${\bf x}=[x_1,\ldots,x_n]$} defined over an $n$-dimensional unit hypercube $\mathbb{R}^n$, is expressed as a linear combination of component functions, as follows:
\begin{eqnarray}
f(\textcolor{black}{{\bf x}}) &=& f_0+
\sum_{i=1}^n f_i(x_i)+ 
\sum_{\genfrac{}{}{0pt}{}{i,j=1}{i<j}}^n 
f_{ij}(x_{i},x_{j})  \nonumber \\ 
&&+ \cdots + f_{1\ldots n}(x_1,\ldots,x_n),
\label{hdmrUzun}
\end{eqnarray}
where $f_0$ is a zero-order component function (constant term) calculated by taking the mean response of \textcolor{black}{$f({\bf x})$}; 
the first order terms $f_{i}$ correspond to individual effects of $x_{i}-$$th$ variable over the model output \textcolor{black}{$f({\bf x})$}; 
the second order component functions $f_{ij}$ refer to the cooperative (crossed) effects of the variables $x_i$ and $x_j$ on the \textcolor{black}{$f({\bf x})$} 
on the model output; the higher order terms correspond to cooperative effects of their associated variables on the model output; and $f_{1\ldots n}(x_1,\ldots,x_n)$ is the residuals holding the $n-$$th$ order dependence of all the variables over \textcolor{black}{$f({\bf x})$}. 
For high dimensional systems, higher order interactions are expected to be weak, and hence Eq. \eqref{hdmrUzun} can be approximately represented as follows: 
\begin{equation}
f(\mathbf{x}) \approx f_0+ \!\!\sum_{i=1}^n f_i(x_i).
\label{hdmrKisa}
\end{equation}
\textcolor{black}{The} computation of the HDMR component functions involves costly multi-dimensional integrations. This is due to the fact that sampling effort required for Monte Carlo integration scales exponentially as the number of variables increases \cite{Alis2001}.  To overcome this difficulty, the first order terms are approximated as a linear combination of some orthogonal base functions as follows: 
\begin{equation}
 f_i(x_i) \approx \sum_{j=1}^p\alpha_{ij}\varphi_j(x_i),
\label{eq:polyapprox}
\end{equation}
where $p$ is the order of orthogonal base functions with corresponding coefficients ${\alpha_{ij}}$ that need to be calculated. Legendre and Chebyshev polynomials are the most common base functions used in the literature.

As mentioned above, all the component functions need to be calculated for an exact representation, but this is a very computationally inefficient process, especially as the size of input space increases, leading to the curse of dimensionality. Despite this, the HDMR is a very efficient and accurate model and works quite well with relatively a few component functions where higher-order interactions between the features are weak \cite{Taskin2019}. Owing to the main assumption in manifold learning, a nonlinear manifold structure underlies the high dimensional space, meaning that there exists a manifold of lower dimensionality than the original space that captures the geometric structure of the data. Therefore, one can claim that the HDMR meets the principle of manifold learning when modeling the transformation between high dimensional and the lower-dimensional latent space. In the proposed method, the optimization problem in  GE is modified by replacing the low-dimensional coordinates of the samples with their HDMR representations, and the optimization problem is readily solved. As a result, both low dimensional coordinates of the samples in high dimensional spaces and the related projection matrix for out-of-samples are obtained.

{\color{black}
\section{Graph Embedding}
Graph embedding (GE) constitutes a general dimensionality reduction framework to describe a wide variety of methods. The GE embeds the high-dimensional data into a low-dimensional space by preserving local neighboring relationships of the high-dimensional data. The neighborhood information is determined based on the similarities between the samples, which is denoted by an adjacent matrix $\mathbf{W}$. The following optimization problem needs to be solved to find the coordinates of the samples in the low dimensional space:
\begin{eqnarray}
y^* &=& \sum_{i,j}(y_i - y_j)^2 W_{ij} =  2 \mathbf{y^\top}\mathbf{L y} \nonumber \\
&=& \underset{{\bf y}^\top {\bf D y} = 1}{\operatorname{arg min}\;} \mathbf{y^\top}\mathbf{L y}  = {\operatorname{arg min}\;} \frac {\mathbf{y^\top}\mathbf{L y}} {\mathbf{y^\top}\mathbf{D y}}, 
\label{eq:geMin}
\end{eqnarray}
where $\mathbf{L}$ is the Laplacian matrix, which is defined as $\bf L= D - W$. The degree matrix $\bf D$ is a diagonal matrix whose elements are column or row sums of $\bf W$, i.e., ${\bf D}_{ii} = \sum_j {W}_{ji}$. The solution of the optimization problem in Eq. (\ref{eq:geMin}) yields the following generalized eigenvalue problem: 
\begin{eqnarray}
{\bf L y} = \lambda {\bf Dy},
\label{eq:graphE}
\end{eqnarray}
where $\mathbf{y}$'s are the eigenvectors associated with the smallest eigenvalues of Eq. \eqref{eq:graphE}, excluding the one that corresponds to the zero eigenvalue.
}

\section{HDMR-Based Manifold Learning}

\subsection{Notation, problem definition and solution}
\textcolor{black}{Throughout this section, matrices and vectors are denoted by bold fonts, with uppercase for matrices and lower case for vectors. The element $(i,j)$ in the matrix (e.g., $\mathbf{A}$) is denoted by ${A}_{ij}$ whereas its column vector is denoted by $\mathbf{a}_{i}$, and we indicate vector concatenation as $\mathbf{A} = [\mathbf{a}_{1}|\cdots|\mathbf{a}_{m}]$, and transposition of vector $\mathbf{a}$ with $\mathbf{a}^\top$.}

\textcolor{black}{Let ${\mathcal X}=\{\mathbf{x}_i \in \mathbb{R}^n\}_{i=1}^m$ be} a set of $m$ samples assumed to lie on (or close to) a manifold embedded in a high-dimensional space. A unified graph embedding framework \textcolor{black}{minimizes} 
\begin{equation}
\textcolor{black}{{\cal J} = \sum_{i,j}^m(y_i - y_j)^2 \textcolor{black}{{W}_{ij}},}
\label{eq:ge}
\end{equation}
where $\textcolor{black}{{W}_{ij}}$ describes the similarity between $\mathbf{x}_i$ and $\mathbf{x}_j$, and $y_i \in \mathbb{R}$ is the mapping of $\mathbf{x}_i$ onto a one-dimensional space. If $\mathbf{x}_i$ and $\mathbf{x}_j$ are nearby points, then $y_i$ and $y_j$ are nearby as well. It is reasonable to assume that the embedded coordinates, $y_i$'s, are the result of an unknown embedding function $f:\mathbb{R}^n \rightarrow \mathbb{R}$. Many different representation schemes can be introduced for describing $f$. Among them, the HDMR seems a promising one because of its ability to divide the individual effects of independent variables on the output. 
\textcolor{black}{
To further simplify the notation, let us also use only the first order terms for an arbitrary point $\mathbf{x}$ as follows: 
\begin{align}
y &= f(x_1, x_2, \ldots, x_n) \approx f_0+ f_1(x_1) + \ldots + f_n(x_n),
\label{eq:hdmrge}
\end{align}
where subscripts indicate the n-$th$ coordinate.
First order HDMR terms $f_k$ can also be approximated by a linear combination of orthonormal polynomials, $\varphi$, as in Eq. \eqref{eq:polyapprox}. 
Let us now collectively group in vectors all the basis functions and the corresponding weights, respectively:
\begin{align}
\myphi & =   \left[ 
  \varphi_{1}(x_1),  \ldots, \varphi_{p}(x_1), \ldots,  \varphi_{1}(x_n),  \ldots, \varphi_{p}(x_n) \right]^\top \in \mathbb{R}^{np}, \nonumber\\
\ba & =   \left[
  {\alpha}_{11},  \ldots, \alpha_{1p},   \ldots,  \alpha_{n1},  \ldots, \alpha_{np}  \right]^\top \in \mathbb{R}^{np} \nonumber 
\end{align}
then, for a particular point ${\bf x}_i$, Eq. \eqref{eq:hdmrge} becomes 
\begin{equation}
    y_i = f_0 + \ba^\top \myphi_i.
    \label{eq:yis}
\end{equation}
Now, substituting Eq. \eqref{eq:yis} into Eq. \eqref{eq:ge} yields the minimization functional
\begin{equation}
{\cal J}  = \sum_{i,j}^m(\ba^\top \myphi_i - \ba^\top \myphi_j)^2 {W}_{ij}.
\label{eq:hdmrgeopt}
\end{equation}
Let us now define the Laplacian matrix as ${\bf L}={\bf D}-{\bf W}$, where ${\bf W}$ is the weight matrix which is an $m\times m$ symmetric matrix, and $\bf D$ is a degree matrix with entries ${D}_{ii} \equiv \sum_{j}{W}_{ij}$. Then, after some simple linear algebra, one can express the minimizing functional ${\cal J}$ in quadratic form as follows:
\begin{align}
{\cal J} &= 2[\ba^\top \myphi_1 | \ldots | \ba^\top \myphi_m] {\bf L} [\ba^\top \myphi_1| \ldots | \ba^\top \myphi_m ]^\top \nonumber \\
&= 2\ba^\top [\myphi_1| \cdots| \myphi_m] {\bf L} [\myphi_1| \cdots| \myphi_m]^\top \ba \nonumber \\
&= 2 {\boldsymbol \ba}^\top \boldsymbol\Phi{\bf L} \boldsymbol\Phi^\top  \ba, \label{eq:finalopt}
\end{align}
where $\boldsymbol\Phi=[\myphi_1 \ldots \myphi_m]$ is an $np \times m$ matrix.}
Minimizing Eq. \eqref{eq:finalopt} poses a trivial solution, i.e., $\ba = \bf 0$, \textcolor{black}{which can be solved by adding a regularization term on the Legendre polynomials, and thus the problem becomes:} 
\begin{align}
\underset{ \ba^\top \boldsymbol\Phi {\bf D} \boldsymbol\Phi^\top \ba  = { 1}}{\operatorname{arg min}\;} 
\ba^\top \boldsymbol\Phi {\bf L} \boldsymbol\Phi^\top \ba  + \beta\|\ba\|^2.
\label{eq:geMin}
\end{align}
The Lagrange functional is then written as follows: 
\begin{eqnarray}
  \mathcal{L}(\ba,\lambda,\beta) = \ba^\top \boldsymbol\Phi {\bf L} \boldsymbol\Phi^\top\ba - \lambda(\ba^\top \boldsymbol\Phi {\bf D}\boldsymbol\Phi^\top \ba - 1) + \beta \ba^\top \ba,
\end{eqnarray}
\textcolor{black}{whose minimization leads to 
\begin{align}
(\boldsymbol\Phi {\bf L} \boldsymbol\Phi^\top + \beta {\bf I}_{np} )\ba & = \lambda \boldsymbol\Phi {\mathbf D}\boldsymbol\Phi^\top \ba,
\label{eq:eigeq}
\end{align}
where $\beta>0$ is an arbitrary regularization term,} and ${\bf I}_{np}$ is an $np\times np$ identity matrix.  The coefficients of LPs are the eigenvectors corresponding to the smallest eigenvalues of Eq. \eqref{eq:eigeq}, excluding the one that corresponds to  zero eigenvalue.  Each  dimension  in  subspace is formed with the subsequent eigenvectors. Note that connections  to  other  semi-supervised  manifold  learning  algorithms presented in the literature can be pointed out \cite{tuia2014unsupervised,tuia2016kernel}.


\subsection{On the basis functions and nonlinear extensions}
Legendre Polynomials (LPs) used as basis functions in the approximation of the HDMR component functions are orthogonal on the domain $[-1,1]$. However, in real-world problems, the distribution of data may not satisfy this assumption. This requires a suitable pre-processing step that essentially rescales LPs on a general interval, i.e., $[a, b]$. It should be noted that the original domain can also be used directly but, in this case, each feature should be scaled a priori, which might distort the existing manifold.

Let us  assume that $P_n(x)$ is the $n-$$th$ Legendre polynomial defined on $[-1,1]$. They are orthogonal satisfying following condition:
\begin{equation}
\int_{-1}^1 P_n(x) P_m(x) dx  = \frac 2 {2n+1} \delta_{n,m}
\label{eq:LPorthogonality}
\end{equation}
where $\delta$ is the Kronecker symbol. We are looking for orthonormal polynomials $\myphi_n(x) = w_n P_n(y(x))$ satisfying 
\begin{equation}
\int_a^b \varphi_n(x) \varphi_m(x) = \delta_{n,m} \label{orthonormal}  
\end{equation}
By changing of the variables $y(x) = 2(x-b)/(b-a)  + 1 $, the weights $w_n$ are determined as follows:
\begin{align}
\delta_{n,m} & = \int_a^b \varphi_n(x) \varphi_m(x) \mathrm{d}x \nonumber \\ 
             & = \int_a^b w_n P_n(y(x)) w_m P_m(y(x)) \mathrm{d}x \nonumber \\ 
            & = w_nw_m\frac{b-a}{2}\int_{-1}^{+1} P_n(y)P_m(y)\mathrm{d}y \\
             & = w_nw_m\frac{b-a}{2n+1}\delta_{n,m}
\label{eq:integralsLP}
\end{align}
Hence $w_n = ((2n+1)/(b-a))^{1/2}.$
Therefore, the general form of the  shifted and scaled Legendre polynomials on $[a,b]$ is:
\begin{equation}
\varphi_n(x) = \left(\frac{2n+1}{b-a}\right)^{1/2} P_n(y(x))
\end{equation}

Another important remark we want to highlight is the certain possibility to suggest nonlinear versions of the proposed method. \textcolor{black}{Among the several ways, we observe that are based on linear algebra, then one could try to `kernelize' the algorithm in a similar way to standard multivariate kernel machines like KPCA or KPLS \cite{Rojo18dspkm}[ch. 14] and KEMA \cite{tuia2016kernel}. The problem with a straightforward kernelization is that one can hardly account for the cross-terms of the mapped functions $f_{ij}$ in a kernel function, which makes the derivation of a kernel function a challenging problem.} This is why we suggested here following an explicit mapping as a more convenient way.

\section{Experiments and Discussion}

\subsection{Dataset Description}
Three hyperspectral datasets, namely, Botswana, the Loukia, and the Indian Pines, are used to show the effectiveness of the proposed method. Each hyperspectral image along with its ground reference information with class legends is given in Fig. \ref{fig:allThematicMaps}. 

      The Botswana image was collected over the  Okavango Delta by the Hyperion sensor on the NASA EO-1 satellite in 2001. It consists of 1476$\times$256 pixels.  After removing the noisy bands, 145 spectral bands remained at 30-m spatial resolution and 10-nm spectral resolution for 14 classes of interest. Three of these classes,  {\it riparian} (class 6) and {\it acacia-woodlands} (class 9), were reported as the most difficult classes because of their high spectral similarities \cite{Ma2015}.
      
     The Indian Pines dataset was collected over the agricultural site in northwestern Indiana by the NASA AVIRIS sensor. Twenty water absorption bands were removed, resulting in 200 spectral bands to be used in the experiments. It consists of $145\times145$ pixels with 20-m spatial resolution and 10-nm spectral resolution. The sixteen land cover classes are considered in this study, two of which reported as the difficult classes are {\it corn-min-tillage} (class 3) and {\it soybeans-clean-tillage} (class 12) \cite{Crawford2011}.
 
     The 
HyRANK-Loukia dataset was obtained using the Hyperion Earth Observing-1 sensor \cite{karantzalos}. It has a spatial resolution of 30 m and spectral coverage from 400 to 2500 nm. Following a pre-processing step, the image provided 176 spectral bands with a pixel size of $249\times945$. 

\subsection{Experimental Setup}
We learn the latent space of each hyperspectral dataset using DR methods on the training set, then the test set (out-of-samples) is projected to the corresponding subspace on which the performance of each DR method is evaluated in terms of classification accuracy. We also include in the comparison the case where all the original spectral features are used for the classification, termed as `Baseline' (BSL). 
As for the comparison of the performance of our method, we used five DR methods  \textcolor{black}{including both linear and nonlinear, as well as supervised and unsupervised versions as follows:}
\begin{itemize}
    \item Locality Preserving Projection (LPP), linear and  unsupervised  \cite{He2004},
    \item Locality Sensitive Discriminant Analysis (LSDA), linear and  supervised \cite{lsda},
    \item Kernel Partial Least Squares Regression (KPLS), nonlinear and  supervised \cite{kpls},
    \item Kernel Orthonormalized PLS (KOPLS), nonlinear and  supervised \cite{kopls}, 
    \item Locality Linear Embedding (LLE), nonlinear and unsupervised \cite{Roweis2000}.
\end{itemize}
{\color{black} It should be noted that we intentionally selected these methods not only because they are methodology-related, but also because several of those DR approaches utilized in comparisons have the potential to theoretically extend other methodologies (e.g. KOPLS over PCA, PLS, OPLS, and their sparse versions) \cite{arenas2008efficient,Rojo18dspkm,ArenasGarcia201316}. Moreover, the use of other methods based on deep learning in the comparison would turn complicated, as they exploit the spatial autocorrelation of the images by construction and do not exploit the graph embedding concept.}

Each DR method used in the comparisons has its projection matrix to map the out-of-samples to the latent space except LLE in which the OOS is projected via the {\em Nystr\"om} approach \cite{Rojo18dspkm,hidalgo2020efficient}. The interested reader can find a comprehensive review of dimensionality reduction methods for remote sensing in \cite{CampsValls09wiley,Rojo18dspkm,ArenasGarcia201316}. 
\textcolor{black}{We used the {\it radial basis function} (RBF) kernel function for KPLS and KOPLS, as it only requires selecting one parameter and has shown good performance in previous works}. 
All the data was normalized to zero-mean and unit variance, and the kernel width is set as the median of all pairwise distances between samples \cite{ArenasGarcia201316}. As the methods, which are LLE, LPP, LSDA, and our method, are graph-based methods,  they use a graph affinity matrix. {\color {black} The LSDA generates the graph matrix with the combination of the within and between class matrices since it is designed as a supervised method. } As known, the performance of the graph-based methods strongly relies on the construction of the graph affinity matrix, depending on two parameters: (i) the weights of the connected samples and (ii) the neighborhood size needed to build the graph. To reduce the effect of the graph affinity matrix on the embedding results, the weights of connected samples are constructed based on the binary relations, meaning that they are set as $1$ if vertices $i$ and $j$ are connected by an edge and are $0$ otherwise. Here, with prior class label information, the samples were connected in a supervised fashion such that an edge is constructed between the $i-$th and $j-$th samples if and only if $x_i-$th and $x_j-$th samples belong to the same class.  {\color {black} As we define the affinity matrix in a supervised fashion, the neighborhood parameter is set with a $5$-fold cross-validation in such a way that it cannot be larger than the number of samples of the corresponding class.} {\color {black} Nevertheless, from our practical experience we did not observe a dramatic change in results when a regular average distance rule of thumb criterion was selected.} Note that the LLE computes the weights of the graph affinity matrix by optimally preserving the local manifold structure between each sample and its neighbors in the high-dimensional space. The parameters of the proposed method, which are the regularization term and the order of the polynomial, are determined based on 5-fold cross-validations for each HSI dataset. 

As classification methods, 1-nn and SVM with RBF kernel are used. The reason for selecting 1-nn in the experiments is because it is naturally coherent with the main idea of manifold learning and does not require any parameter setting as well. \textcolor{black}{We used SVM to measure the degree of the separability of the features (`expressiveness') in the embedding space as well as to measure the dependency on the classifier.} We empirically tuned the parameters of the SVM for each HSI dataset and used the same settings in the evaluation of all the DR methods for a fair comparison.
A total of 10$\%$ of ground truth was randomly selected as training samples for each hyperspectral dataset, and a total of 10 runs with different random sample selections was used. The results are averaged in all the experiments. The number of training and test samples for each class on three HSI datasets \textcolor{black}{are given in Table \ref{table:trntst}. }

\begin{table}[!h]
\scriptsize
\centering
\caption{Number of training and test samples for each class of HSI datasets used in the experiments.} 
\begin{tabular}{
>{}p{0.04\textwidth}|
>{}p{0.04\textwidth}
>{}p{0.04\textwidth}|
>{}p{0.04\textwidth}
>{}l |
>{}l 
>{}l }
\hline
      & \multicolumn{2}{l|}{\bf Botswana} & \multicolumn{2}{l|}{\bf Loukia} & \multicolumn{2}{l}{\bf Indian Pines} \\
\bf Class & \bf Train                     & \bf Test                     & \bf Train                    & \bf Test                    & \bf Train                       & \bf Test                       \\
\hline 
1     & 27                        & 243                      & 15                       & 129                     & 5                           & 41                         \\
2     & 11                        & 90                       & 4                        & 30                      & 143                         & 1285                       \\
3     & 26                        & 225                      & 28                       & 243                     & 83                          & 747                        \\
4     & 22                        & 193                      & 4                        & 36                      & 24                          & 213                        \\
5     & 27                        & 242                      & 71                       & 630                     & 49                          & 434                        \\
6     & 27                        & 242                      & 12                       & 100                     & 73                          & 657                        \\
7     & 26                        & 233                      & 25                       & 225                     & 3                           & 25                         \\
8     & 21                        & 182                      & 54                       & 482                     & 48                          & 430                        \\
9     & 32                        & 282                      & 190                      & 1707                    & 2                           & 18                         \\
10    & 25                        & 223                      & 141                      & 1261                    & 98                          & 874                        \\
11    & 31                        & 274                      & 21                       & 181                     & 246                         & 2209                       \\
12    & 19                        & 162                      & 25                       & 219                     & 60                          & 533                        \\
13    & 27                        & 241                      & 70                       & 627                     & 21                          & 184                        \\
14    & 10                        & 85                       & 23                       & 203                     & 127                         & 1138                       \\
15    & -                       & -                     & -                      & -                    & 39                          & 347                        \\
16    & -                           &   -                       &     -                     &    -                     & 10                          & 83                         \\
\hline  
\bf Total & \bf 331                       & \bf 2917                     & \bf 683                     & \bf 6073                   & \bf 1031                        & \bf 9218                      \\
\hline
\end{tabular}
\label{table:trntst}
\end{table}


{\color{black} The feature expressiveness in the latent space of the proposed method is also analyzed in terms of three criteria: an information-theoretic estimate of Normalized Mutual Information (NMI) \cite{schutze2008introduction}, Fisher’s Score (FC) \cite{anderson1962introduction,Taskin2019}, and the silhouette’s score (SC) \cite{KaufmanR90}. The SC is a measure of the level of agreement of a clustering result with the given class information while the NMI measures similarity using mutual information. For clustering, $K$-means is used in this study, and the number of $K$ is set to the number of classes of the associated HSI dataset. The FC measures the separability of the classes using within- and between-class matrices.}

\subsection{Results}
To demonstrate the performance of the proposed method, we conducted the following experiments.
\subsubsection{Sensitivity of the Proposed Method to its Parameters}
\begin{figure}[!h]
\centering
\includegraphics[width=3.5in]{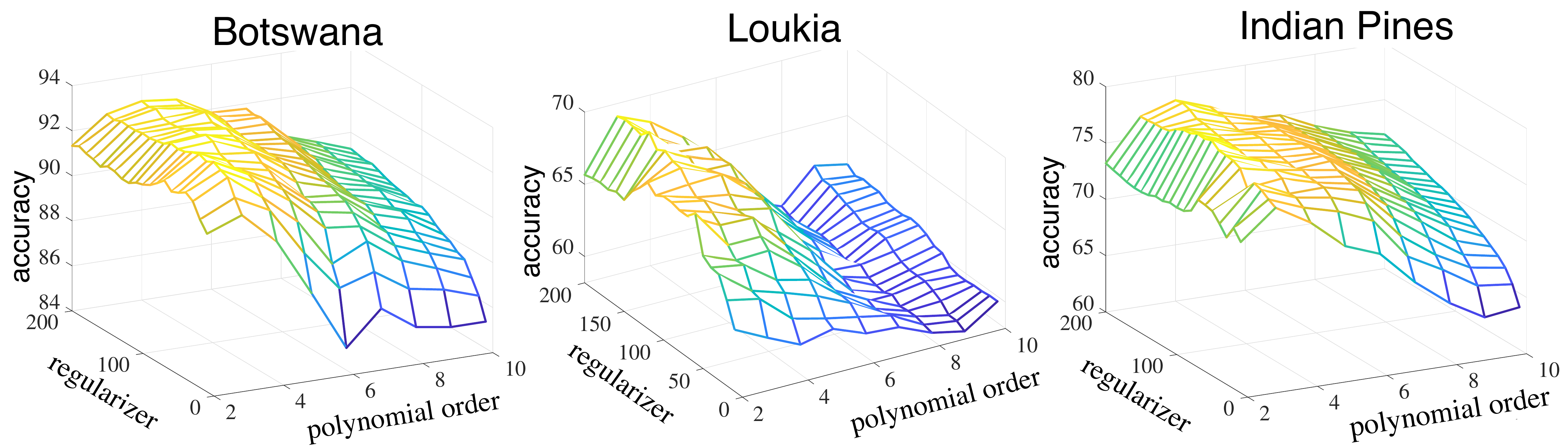} 
\caption{Parameter analysis of HDMR based manifold learning in terms of 1-nn overall classification accuracy for each HSI.}
\label{fig:polyEffects}
\end{figure}

 The parameters of the HDMR-based manifold learning are analyzed in terms of classification accuracy with a grid-search strategy based on $5$-fold cross-validation. We only show the results with the 1-nn classifier in Fig. \ref{fig:polyEffects} (as similar results were obtained with the SVM classifier, results were not shown). The intervals for the regularizer term, $\beta$, and the order of the polynomial, $p$, were determined as $[0, 10, \ldots, 200]$ and $[2, 3, \ldots, 10]$, respectively.

Figure \ref{fig:polyEffects} shows that, as we increase the order of polynomials, higher performance is obtained, but then the regularization parameter plays an important role to alleviate overfitting. Therefore, a higher value of the regularizer term is needed as the order of the polynomial (model complexity) increases, but the regularizer no longer has any effect on the accuracy when the order of the polynomial is selected between about 4 and 10. According to these results, the parameters of the proposed method are finally determined as $p=[4, 3, 4]$ and $\beta =[100, 150, 150]$ for Botswana, Loukia, and Indian Pines, respectively. This parameter configuration was used for the rest of the analysis.


\subsubsection{Sensitivity to Reduced Dimensionality}

\begin{figure}[!h]
\centering
\includegraphics[width=3.5in]{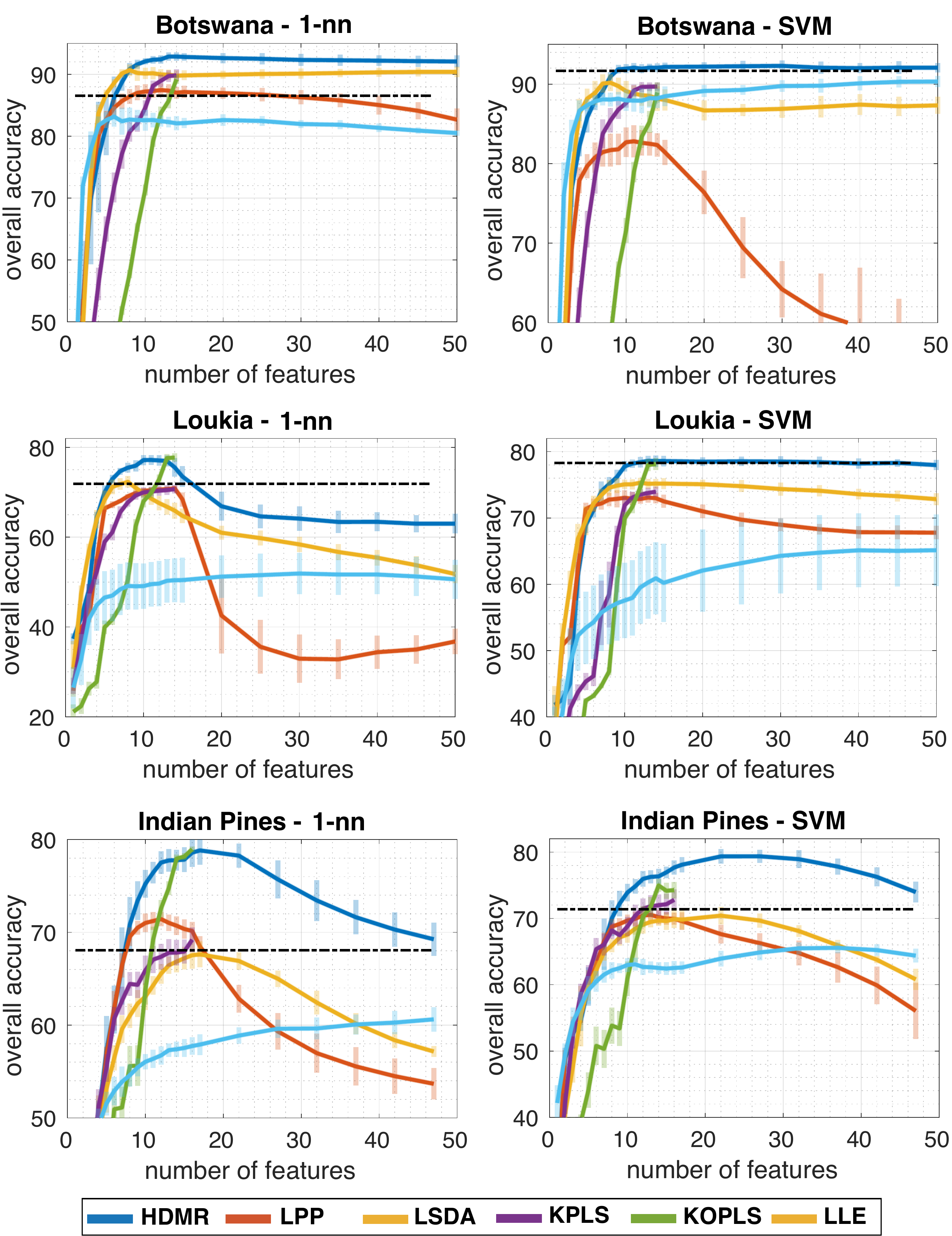} 
\caption{Learning curves of averaged overall accuracy over ten experiments for Botswana, Loukia, and Indian Pines datasets. The dashed black line refers to the case, called {\it Baseline},  when all the features in the ambient space is used in the classification.}
\label{fig:learningrates}
\end{figure}

Fig. \ref{fig:learningrates} shows the learning curves of each DR method as a function of the size of the embedding space, which is fixed to 50. Note that KPLS and KOPLS \textcolor{black}{can provide as many features in the low-dimensional space as the number of classes} of the associated HSI dataset. The classification using all the spectral features was also conducted to show the performance of the DR methods. Both 1-nn and SVM classifiers were used to evaluate the quality of the embedding space. {\color{black} The classification accuracy increases in agreement with the number of features in the embedding space, until it reaches a point where adding more features minimally affects classification performance. This could be due to the fact that the remaining features are not as discriminative as the first extracted ones. However, in some experiments, the learning curves behave differently, with a decrease in classification accuracy when more features are added, implying that some of the extracted features are not only redundant but also have a negative impact on 1-nn and SVM.} The results show that the proposed method has superior performance compared to the other methods, and is also much better than the {\it Baseline} case. The HDMR also performs the best independently of the classifier, meaning that the extracted features in the embedding space are more expressive 
than those of other methods. The highest classification performance (averaged on the overall classification accuracy) along with the Kappa values, $\kappa$ that are achieved for each HSI dataset in the embedding space with the dimension, $d$, are provided for both classifiers in Table \ref{table:learningratetable}.

\begin{table}[!h]
\centering
\caption{Best OA classification accuracy, Kappa values, and size of dimension of  the embedding space achieved at the embedding space.}
\begin{tabular} {p{0.055\textwidth}p{0.01\textwidth}p{0.03\textwidth}p{0.04\textwidth}p{0.04\textwidth}p{0.03\textwidth}p{0.03\textwidth}p{0.02\textwidth}p{0.03\textwidth}}
\hline 
 &               & \bf BSL & \bf HDMR & \bf KOPLS & \bf KPLS & \bf LSDA &\bf  LPP &\bf  LLE \\ 
\hline 
\bf Botswana    & OA        & 86.9   & \bf  93.0   & 88.1   & 89.2  & 89.8  & 84.7  & 84.2   \\

                   & $\kappa$     & 85.8   & \bf  92.5   & 87.1   & 88.3  & 89.0  & 83.4  & 82.9   \\
1-nn
             & $d$ & 145    & 15     & 14     & 14    & 8     & 13    & 7      \\
\rowcolor[HTML]{ECF4FF}
\cline{2-9} \rowcolor[HTML]{ECF4FF}
\cellcolor[HTML]{ECF4FF}             & OA        & 92.2  & \bf  92.5   & 88.1   & 90.5  & 90.2  & 82.8  & 88.4   \\
\rowcolor[HTML]{ECF4FF}
\cellcolor[HTML]{ECF4FF}                 & $\kappa$     & 91.2   & \bf  91.9   & 87.1   & 89.7  & 89.1  & 81.4  & 87.4   \\
\rowcolor[HTML]{ECF4FF}
\cellcolor[HTML]{ECF4FF} SVM          & $d$ & 145    & 13     & 14     & 14    & 8     & 10    & 18     \\ 
\hline 
\bf Loukia   & OA        & 70.4   & \bf  78.4   & 77.7   & 70.5  & 71.2  & 70.6  & 50.7   \\

                 & $\kappa$     & 64.9   & \bf 74.4   & 73.3   & 65.0  & 69.5  & 66.9  & 42.2   \\
 1-nn
                  & $d$ & 176    & 13     & 14     & 14    & 11    & 14    & 12      \\
\cline{2-9} \rowcolor[HTML]{ECF4FF}
\rowcolor[HTML]{ECF4FF}
\cellcolor[HTML]{ECF4FF}              & OA        & 78.9   & \bf  79.7   & 78.5   & 74.2  & 75.1  & 74.6  & 65.3   \\
\rowcolor[HTML]{ECF4FF}
\cellcolor[HTML]{ECF4FF}                   & $\kappa$     & 74.7   & \bf  75.8   & 74.1   & 67.6  & 70.2  & 69.8  & 56.3   \\
\rowcolor[HTML]{ECF4FF} SVM
\cellcolor[HTML]{ECF4FF}                   & $d$ & 176    & 12     & 14     & 14    & 16    & 10    & 40     \\
 
\hline 
\bf Indian  &  OA        & 67.6   & 77.9   & \bf  78.8   & 68.4  & 68.1  & 73.5  & 60.1   \\

  \bf Pines               & $\kappa$     & 63.0   & 74.8   & \bf  75.7   & 64.2  & 65.2  & 69.7  & 53.1   \\
 1-nn
                &  $d$ & 200    & 14     & 16     & 16    & 18    & 13    & 50     \\
\cline{2-9} \rowcolor[HTML]{ECF4FF}
\rowcolor[HTML]{ECF4FF}
\cellcolor[HTML]{ECF4FF}              & OA        & 72.8  & \bf 78.2   & 78.0  & 75.1 & 71.3 & 70.1 & 65.2  \\
\rowcolor[HTML]{ECF4FF}
\cellcolor[HTML]{ECF4FF}                  & $\kappa$     & 68.2  & \bf 75.2   & 74.9  & 71.6 & 67.5 & 66.4 & 59.9  \\
\rowcolor[HTML]{ECF4FF} SVM
\cellcolor[HTML]{ECF4FF}                  & $d$ & 200    & 20     & 16     & 12    & 18    & 12    & 32 \\
\hline
\end{tabular}
\label{table:learningratetable}
\end{table}


\begin{table*}[!h]
\caption{\color{black} Silhouette’s score (SC), information-theoretic estimate of Normalized Mutual Information (NMI), and Fisher’s Score (FC) of  features at the corresponding latent space. }
\centering
\begin{tabular}{lllllllll}  
\hline                               &              & \textbf{Baseline} & \textbf{HDMR}   & \textbf{LPP} & \textbf{LSDA} & \textbf{KPLS} & \textbf{KOPLS} & \textbf{LLE} \\ \hline
\multirow{3}{*}{\textbf{Botswana}}     & \textbf{SC}  & 0.208             & \textbf{0.342}  & 0.187        & 0.214         & 0.34          & 0.305          & 0.182        \\
                                       & \textbf{NMI} & 0.714             & \textbf{0.844}  & 0.83         & 0.759         & 0.811         & 0.823          & 0.701        \\
                                       & \textbf{FC}  & 120               & \textbf{22000}  & 250          & 4900          & 150           & 9.5            & 120          \\ \hline
\multirow{3}{*}{\textbf{Loukia}}       & \textbf{SC}  & 0.048             & \textbf{0.229}  & 0.044        & 0.079         & 0.102         & 0.125          & -0.74        \\
                                       & \textbf{NMI} & 0.018             & \textbf{0.6663} & 0.46         & 0.486         & 0.541         & 0.538          & 0.02         \\
                                       & \textbf{FC}  & -140000           & \textbf{450}    & 25           & 59            & 81            & 58             & -140000      \\ \hline
\multirow{3}{*}{\textbf{Indian Pines}} & \textbf{SC}  & -0.027            & \textbf{0.134}  & 0.049        & -0.016        & 0.026         & 0.127          & -0.067       \\
                                       & \textbf{NMI} & 0.414             & \textbf{0.587}  & 0.448        & 0.458         & 0.478         & 0.577          & 0.424        \\
                                      & \textbf{FC}  & \textbf{480}      & 180             & 24           & 300           & 310           & 130            & 480          \\ \hline
\end{tabular}
\label{table:featureRepresentation}
\end{table*}

\begin{figure*}[!t]
\centering
\includegraphics[width=7in]{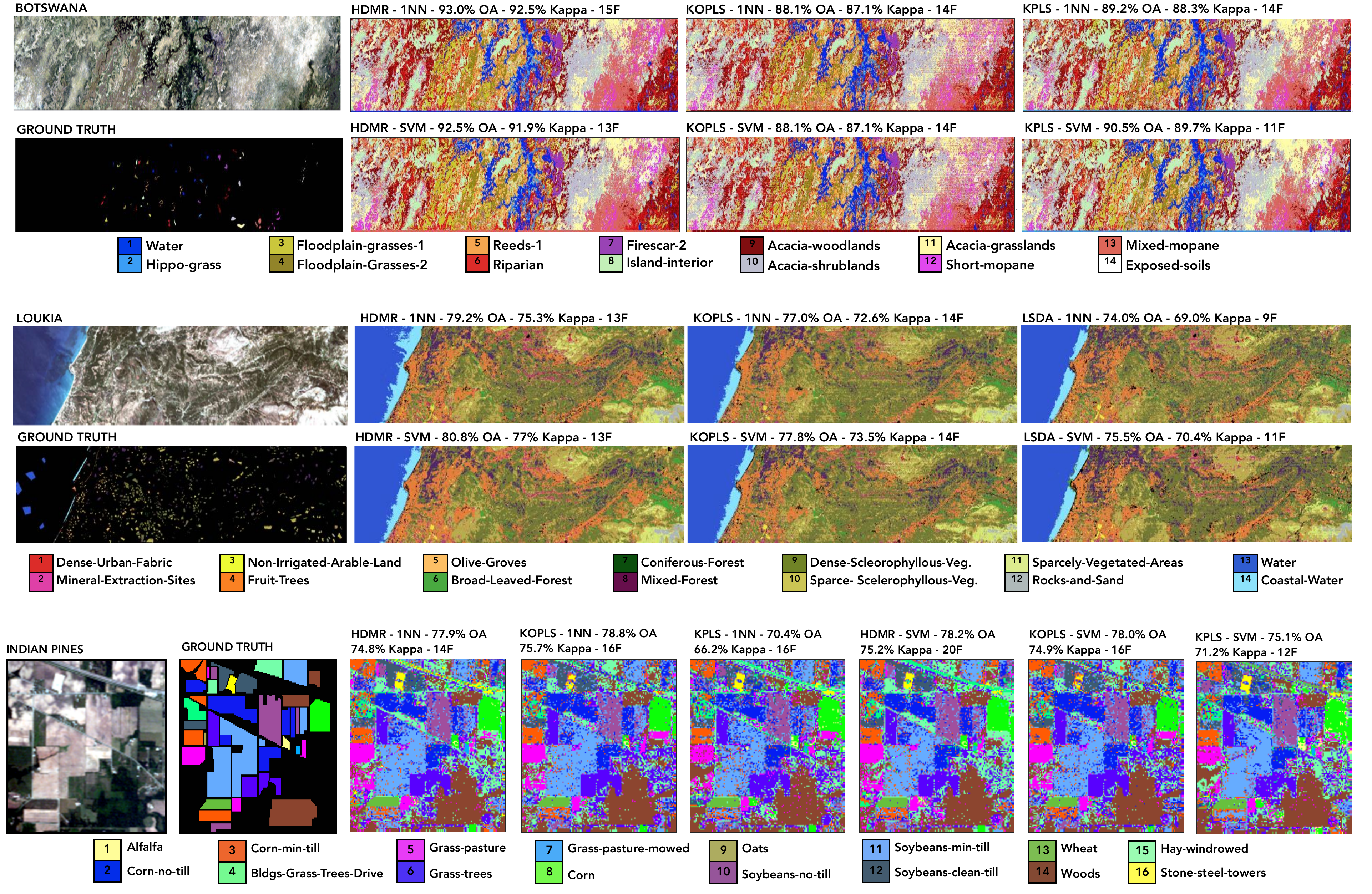} 
\caption{{HSIs along with ground truth data with class label names and classification maps of different DR methods obtained using 1-nn and SVM classifier on the associated HSI dataset.}
\label{fig:allThematicMaps}}
\end{figure*}

\textcolor{black}{
According to the summarizing Table \ref{table:learningratetable}, the HDMR achieves the highest $\kappa$'s agreement value in the Botswana dataset; 92.5\% and 91.9\% for $1$-nn and SVM, respectively, even when using relatively \textcolor{black}{very few features for projection onto} the embedding space. In the Loukia dataset, KOPLS achieves similar performance to the HDMR in terms of $\kappa$ values, approximately 78.0\% and 74.0\% for $1$-nn and SVM, respectively, but it does not perform well \textcolor{black}{when using the top features} 
as observed in Fig \ref{fig:learningrates}. In the Indian Pines, the HDMR significantly outperforms all the methods, except the KOPLS, when using only $14$ features of the embedding space. Note that the HDMR is competitive with KOPLS when using as many features as the number of classes, but it is superior to KOPLS, especially with the first features of the embedding space. In all the experiments, we observed that the LLE performs \textcolor{black}{poorly} 
with a relatively high standard deviation in the overall accuracy. 
The reasons might be that LLE does not use prior class information since it is unsupervised and is highly sensitive to noise, data distribution, variability, as well as multicollinearity, \textcolor{black}{as pointed out in}  \cite{Hong2017,Goldberg}. Moreover, in most cases, the performance of the LPP and LSDA dramatically decreases independent\textcolor{black}{ly} of the classifier as the number of features increases in the embedding space, meaning that the selection of intrinsic dimensionality for these two methods becomes more critical compared to the other methods.}
\textcolor{black}{Figure \ref{fig:allThematicMaps} shows the classification maps of all the HSIs obtained with 1-nn and SVM for visual assessment. Note that the classification maps only for three methods out of all the DR methods, performing the highest performance, were only presented in this paper. The overall accuracy (OA), the $\kappa$ value, and the size of the embedding space, illustrated as e.g., {\it 15F, 14F}, and so on, in which the classification maps are obtained, are given at \textcolor{black}{the} top of each figure. Overall, it can be seen the classification map of the HDMR \textcolor{black}{shows} 
less salt-and-paper \textcolor{black}{classification maps} 
compared to the other methods. \textcolor{black}{This indicates} 
that the HDMR provides more discriminative \textcolor{black}{and robust} features in the embedding space. As for the class-wise accuracy shown in Table \ref{table:classwise}, in the Botswana dataset, the HDMR obtains highly competitive results for each class, particularly for {\it Hippo-Grass, Floodplain-Grass-1, Floodplain-Grass-2, Riparian, Acacia-Woodlands, Acacia-Grasslands}, and {\it Short-Mopane} that have a significant improvement of around 3\% to 10\% in classification accuracy. In the Loukia dataset, the HDMR performs very well  especially for the classes, {\it Non-Irrigated-Arable-Land, Broad-Leaved-Forest, Dense Scleorophyllous-Veg., Sparse-Scleroph Veg.,} and {\it Rocks-and-Sands}, with an improvement of around 3\%-15\% in the accuracy. In the Indian Pines dataset, since the HDMR is competitive with KOPLS in terms of overall accuracy, the class-wise accuracies are similar to each other. However, the HDMR achieves a significant performance in particular for a certain number of classes that are  {\it Grass-Pastured-Mowed, Soybeans-No-Till, Soybeans-Min-Till,  Soybeans-Clean-Till,} and {\it Weat}. Note that the number of training samples for these specific classes is relatively small compared to those of the other classes.}

\begin{table*}[]
\caption{Class-wise accuracy comparison between HDMR and KOPLS in the latent space.}
\centering
\scriptsize
\begin{tabular}{lp{0.04\textwidth}p{0.035\textwidth}p{0.035\textwidth}p{0.035\textwidth}p{0.035\textwidth}p{0.035\textwidth}p{0.035\textwidth}p{0.035\textwidth}p{0.035\textwidth}p{0.035\textwidth}p{0.035\textwidth}p{0.035\textwidth}}
     & \multicolumn{4}{c}{ \bf Botswana}                                                            & \multicolumn{4}{c}{\bf  Loukia}                                                              & \multicolumn{4}{c}{\bf Indian Pines}                                                        \\ \hline
  Class ID     & \multicolumn{2}{c}{\cellcolor[HTML]{ECF4FF}KNN}               & \multicolumn{2}{c}{SVM} & \multicolumn{2}{c}{\cellcolor[HTML]{ECF4FF}KNN}               & \multicolumn{2}{c}{SVM} & \multicolumn{2}{c}{\cellcolor[HTML]{ECF4FF}KNN}               & \multicolumn{2}{c}{SVM} \\
      & \cellcolor[HTML]{ECF4FF}HMDR  & \cellcolor[HTML]{ECF4FF}KOPLS & HMDR  & KOPLS & \cellcolor[HTML]{ECF4FF}HMDR  & \cellcolor[HTML]{ECF4FF}KOPLS & HMDR  & KOPLS & \cellcolor[HTML]{ECF4FF}HMDR  & \cellcolor[HTML]{ECF4FF}KOPLS & HMDR  & KOPLS \\ \hline
1     & \cellcolor[HTML]{ECF4FF}\textbf{100.0} & \cellcolor[HTML]{ECF4FF}99.6           & \textbf{100.0}  & 99.6           & \cellcolor[HTML]{ECF4FF}\textbf{66.7}  & \cellcolor[HTML]{ECF4FF}51.2           & \textbf{66.7}   & 55.0           & \cellcolor[HTML]{ECF4FF}14.6          & \cellcolor[HTML]{ECF4FF}\textbf{22.0}  & 14.6           & \textbf{19.5}   \\
2     & \cellcolor[HTML]{ECF4FF}\textbf{98.9}  & \cellcolor[HTML]{ECF4FF}94.4           & \textbf{98.9}   & 94.4           & \cellcolor[HTML]{ECF4FF}\textbf{73.3}  & \cellcolor[HTML]{ECF4FF}43.3           & 53.3            & \textbf{66.7}  & \cellcolor[HTML]{ECF4FF}70.3          & \cellcolor[HTML]{ECF4FF}\textbf{75.2}  & 73.8           & \textbf{77.0}   \\
3     & \cellcolor[HTML]{ECF4FF}\textbf{97.3}  & \cellcolor[HTML]{ECF4FF}91.6           & \textbf{93.3}   & 91.6           & \cellcolor[HTML]{ECF4FF}\textbf{87.7}  & \cellcolor[HTML]{ECF4FF}72.4           & \textbf{84.8}   & 74.9           & \cellcolor[HTML]{ECF4FF}65.1          & \cellcolor[HTML]{ECF4FF}\textbf{70.3}  & 72.2           & \textbf{72.4}   \\
4     & \cellcolor[HTML]{ECF4FF}94.8           & \cellcolor[HTML]{ECF4FF}\textbf{96.4}  & \textbf{97.9}   & 96.4           & \cellcolor[HTML]{ECF4FF}2.8            & \cellcolor[HTML]{ECF4FF}\textbf{22.2}  & 8.3             & \textbf{11.1}  & \cellcolor[HTML]{ECF4FF}\textbf{40.4} & \cellcolor[HTML]{ECF4FF}32.4           & \textbf{37.1}  & 30.1            \\
5     & \cellcolor[HTML]{ECF4FF}\textbf{81.4}  & \cellcolor[HTML]{ECF4FF}73.6           & \textbf{86.4}   & 73.6           & \cellcolor[HTML]{ECF4FF}\textbf{87.0}  & \cellcolor[HTML]{ECF4FF}86.7           & 86.2            & \textbf{86.5}  & \cellcolor[HTML]{ECF4FF}89.2          & \cellcolor[HTML]{ECF4FF}\textbf{89.6}  & \textbf{90.3}  & 89.2            \\
6     & \cellcolor[HTML]{ECF4FF}\textbf{80.6}  & \cellcolor[HTML]{ECF4FF}78.5           & 76.9            & \textbf{78.5}  & \cellcolor[HTML]{ECF4FF}\textbf{58.0}  & \cellcolor[HTML]{ECF4FF}52.0           & 34.0            & \textbf{50.0}  & \cellcolor[HTML]{ECF4FF}96.8          & \cellcolor[HTML]{ECF4FF}\textbf{98.6}  & 96.4           & \textbf{98.6}   \\
7     & \cellcolor[HTML]{ECF4FF}\textbf{98.7}  & \cellcolor[HTML]{ECF4FF}97.0           & \textbf{99.1}   & 97.0           & \cellcolor[HTML]{ECF4FF}56.4           & \cellcolor[HTML]{ECF4FF}\textbf{61.3}  & 62.7            & \textbf{63.1}  & \cellcolor[HTML]{ECF4FF}\textbf{92.0} & \cellcolor[HTML]{ECF4FF}36.0           & \textbf{44.0}  & 40.0            \\
8     & \cellcolor[HTML]{ECF4FF}\textbf{97.8}  & \cellcolor[HTML]{ECF4FF}95.6           & \textbf{100.0}  & 95.6           & \cellcolor[HTML]{ECF4FF}\textbf{58.5}  & \cellcolor[HTML]{ECF4FF}54.4           & \textbf{67.4}   & 54.1           & \cellcolor[HTML]{ECF4FF}\textbf{97.9} & \cellcolor[HTML]{ECF4FF}96.7           & \textbf{96.7}  & 94.0            \\
9     & \cellcolor[HTML]{ECF4FF}\textbf{86.9}  & \cellcolor[HTML]{ECF4FF}79.1           & \textbf{90.1}   & 79.1           & \cellcolor[HTML]{ECF4FF}73.5           & \cellcolor[HTML]{ECF4FF}\textbf{78.0}  & 79.1            & \textbf{80.1}  & \cellcolor[HTML]{ECF4FF}\textbf{50.0} & \cellcolor[HTML]{ECF4FF}22.2           & \textbf{61.1}  & 22.2            \\
10    & \cellcolor[HTML]{ECF4FF}\textbf{97.3}  & \cellcolor[HTML]{ECF4FF}91.0           & \textbf{95.5}   & 91.0           & \cellcolor[HTML]{ECF4FF}\textbf{82.3}  & \cellcolor[HTML]{ECF4FF}81.7           & \textbf{82.4}   & 81.0           & \cellcolor[HTML]{ECF4FF}\textbf{72.2} & \cellcolor[HTML]{ECF4FF}69.9           & \textbf{72.1}  & 70.7            \\
11    & \cellcolor[HTML]{ECF4FF}\textbf{91.2}  & \cellcolor[HTML]{ECF4FF}89.4           & \textbf{91.6}   & 89.4           & \cellcolor[HTML]{ECF4FF}\textbf{73.5}  & \cellcolor[HTML]{ECF4FF}60.8           & 58.0            & \textbf{60.8}  & \cellcolor[HTML]{ECF4FF}76.3          & \cellcolor[HTML]{ECF4FF}\textbf{78.1}  & \textbf{74.5}  & \textbf{74.5}   \\
12    & \cellcolor[HTML]{ECF4FF}\textbf{95.1}  & \cellcolor[HTML]{ECF4FF}75.9           & \textbf{88.9}   & 75.9           & \cellcolor[HTML]{ECF4FF}\textbf{78.5}  & \cellcolor[HTML]{ECF4FF}71.7           & \textbf{74.4}   & 73.1           & \cellcolor[HTML]{ECF4FF}\textbf{78.4} & \cellcolor[HTML]{ECF4FF}74.9           & \textbf{77.9}  & 76.4            \\
13    & \cellcolor[HTML]{ECF4FF}\textbf{97.1}  & \cellcolor[HTML]{ECF4FF}92.5           & \textbf{95.9}   & 92.5           & \cellcolor[HTML]{ECF4FF}\textbf{100.0} & \cellcolor[HTML]{ECF4FF}\textbf{100.0} & \textbf{100.0}  & \textbf{100.0} & \cellcolor[HTML]{ECF4FF}\textbf{98.4} & \cellcolor[HTML]{ECF4FF}96.7           & 96.2           & \textbf{96.7}   \\
14    & \cellcolor[HTML]{ECF4FF}\textbf{94.1}  & \cellcolor[HTML]{ECF4FF}78.8           & \textbf{78.8}   & \textbf{78.8}  & \cellcolor[HTML]{ECF4FF}\textbf{100.0} & \cellcolor[HTML]{ECF4FF}\textbf{100.0} & \textbf{100.0}  & \textbf{100.0} & \cellcolor[HTML]{ECF4FF}90.5          & \cellcolor[HTML]{ECF4FF}\textbf{91.7}  & 91.0           & \textbf{91.1}   \\
15    & \cellcolor[HTML]{ECF4FF}               & \cellcolor[HTML]{ECF4FF}               &                 &                & \cellcolor[HTML]{ECF4FF}               & \cellcolor[HTML]{ECF4FF}               &                 &                & \cellcolor[HTML]{ECF4FF}60.5          & \cellcolor[HTML]{ECF4FF}\textbf{66.6}  & 62.0           & \textbf{64.8}   \\
16    & \cellcolor[HTML]{ECF4FF}               & \cellcolor[HTML]{ECF4FF}               &                 &                & \cellcolor[HTML]{ECF4FF}               & \cellcolor[HTML]{ECF4FF}               &                 &                & \cellcolor[HTML]{ECF4FF}\textbf{79.5} & \cellcolor[HTML]{ECF4FF}47.0           & \textbf{72.3}  & 32.5            \\ \hline
OA    & \cellcolor[HTML]{ECF4FF}\textbf{93.0}  & \cellcolor[HTML]{ECF4FF}88.1           & \textbf{92.5}   & 88.1           & \cellcolor[HTML]{ECF4FF}\textbf{79.2}  & \cellcolor[HTML]{ECF4FF}77.0           & \textbf{80.8}   & 77.8           & \cellcolor[HTML]    {ECF4FF}77.9          & \cellcolor[HTML]{ECF4FF}\textbf{78.8}  & \textbf{78.3}  & 78.0            \\
$\kappa$ & \cellcolor[HTML]{ECF4FF}\textbf{92.5}  & \cellcolor[HTML]{ECF4FF}87.1           & \textbf{91.9}   & 87.1           & \cellcolor[HTML]{ECF4FF}\textbf{75.3}  & \cellcolor[HTML]{ECF4FF}72.6           & \textbf{77.0}   & 73.5           & \cellcolor[HTML]{ECF4FF}74.8          & \cellcolor[HTML]{ECF4FF}\textbf{75.7}  & 74.8           & \textbf{74.9}   \\ \hline
\end{tabular}
\label{table:classwise}
\end{table*}
{\color{black}
Additionally, we  measure the expressiveness of the features in the latent space with classifier-independent criteria, such as an information-theoretic estimate of Normalized Mutual Information (NMI) \cite{schutze2008introduction}, Fisher’s Score (FC) \cite{anderson1962introduction,Taskin2019}, and the silhouette’s score (SC) \cite{KaufmanR90}. Table \ref{table:featureRepresentation} shows averaged SC, NMI, and FC values for each HSI dataset, with the best values indicated in boldface. The SC ranges from  -1 and 1.  The large positive values indicate strongly separated clustering, whereas negative values indicate that some clustering is mixed. The NMI ranges from 0 to 1, the higher the value, the better the clustering result. Note that the performance of clustering is compared to the class labels when measuring the SCI and NMI. The FC computes a score using within-class and between-class matrices, and the higher the FC value, the more separable the features. According to these results, the features obtained with the proposed method are more relevant than those obtained with the other methods.}









\subsection{Sensitivity to the signal-to-noise (SNR)}

\begin{figure}[!t]
\centering
\includegraphics[width=3.4in]{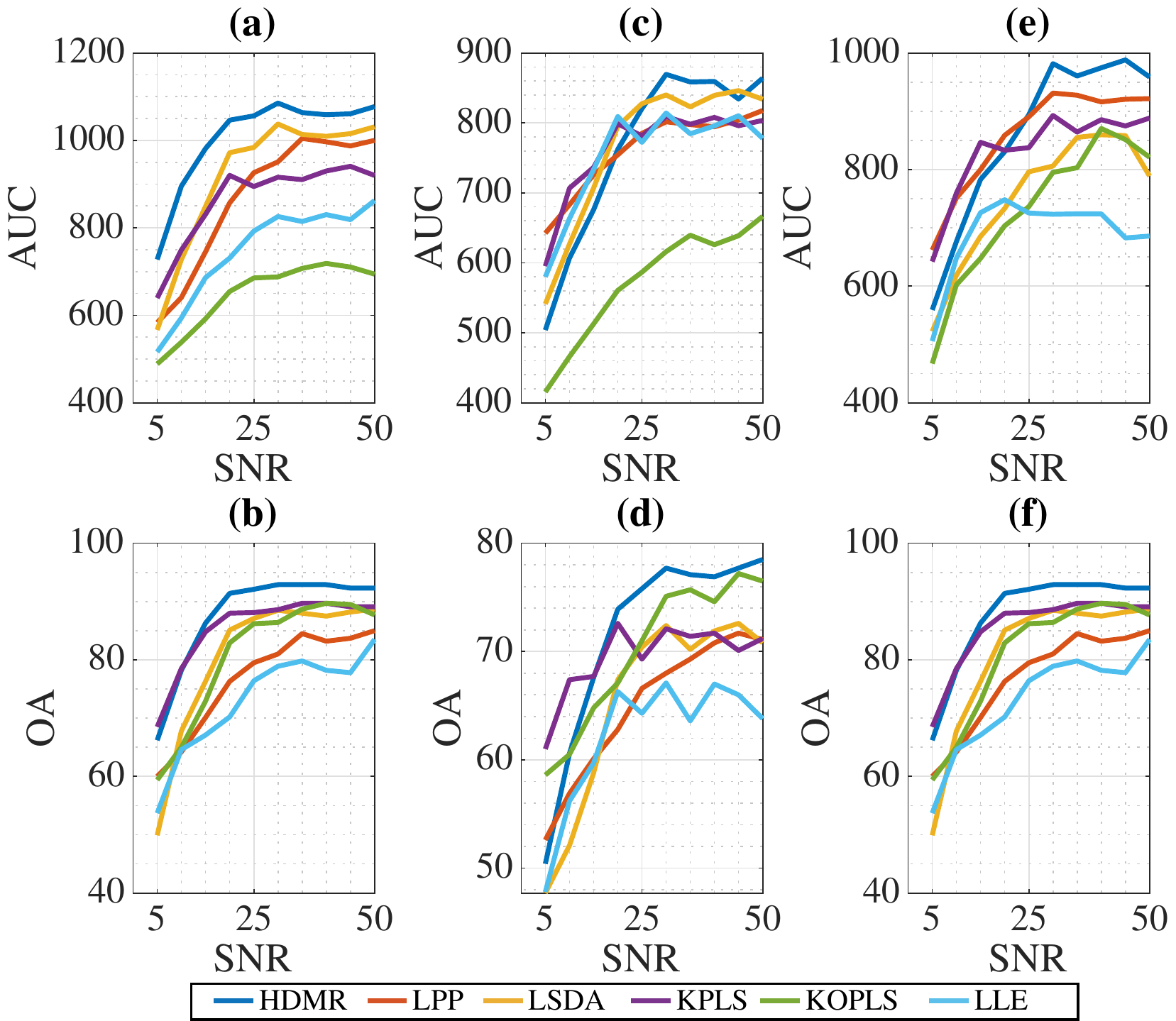} 
\caption{Area under curve and overall accuracy plots for Botswana (a-b), Louika (c-d), and Indian Pines (e-f) when 1-nn is performed.}
\label{fig:snrKNN}
\end{figure}

\begin{figure}[!t]
\centering
\includegraphics[width=3.4in]{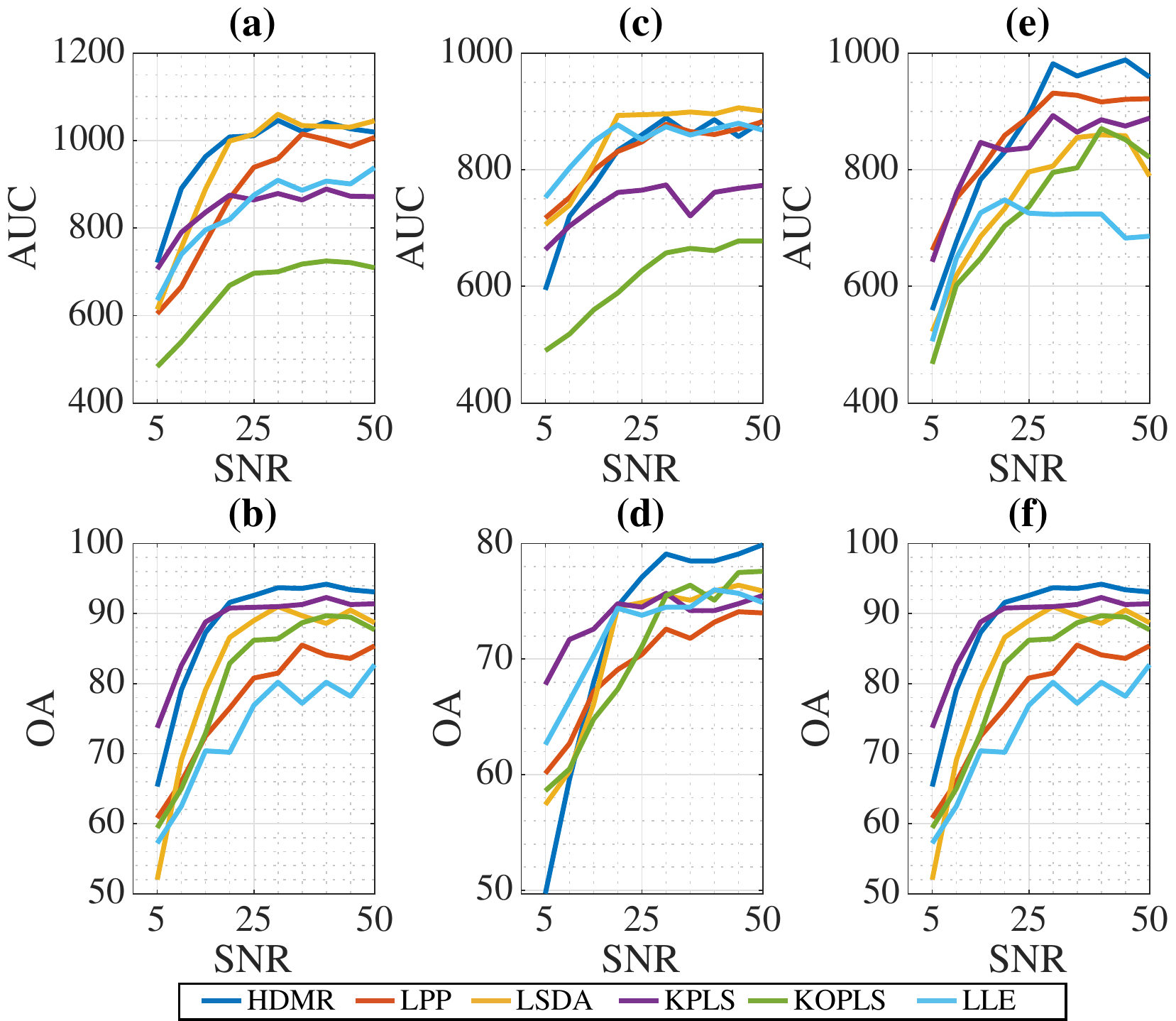} 
\caption{Area under curve and overall accuracy plots for Botswana (a-b), Louika (c-d), and IndianPines (e-f) when SVM is performed.}
\label{fig:snrSVM}
\end{figure}

To analyze the robustness of the methods to the level of noise, the signal-to-noise (SNR) was varied by adding i.i.d. Gaussian noise to ground truth datasets, and the implementations were repeated both for $1$-nn and SVM. The area under curve (AUC) is calculated based on the learning curves of overall accuracy achieved at the embedding space, each of which is constructed with the corresponding manifold learning method. Fig. \ref{fig:snrKNN} and Fig. \ref{fig:snrSVM} show the sensitivity of each DR method to the SNR in terms of learning curves of the AUC and the overall accuracy for 1-nn and SVM classifiers, respectively. 


According to Fig. \ref{fig:snrKNN} (a-b), the HDMR-based ML method is the most robust compared to the rest, providing the highest classification accuracy with the highest AUC at each SNR for the Botswana dataset. For the other HSI datasets, the proposed method shows promising performance in terms of both robustness to SNR and classification accuracy, especially in moderate (yet realistic) noise regimes of SNR values between 25-50 dB. In Fig. \ref{fig:snrKNN}(d), overall accuracy achieved by all the methods varies between 50-68\% when SNR is 5 and 10. In this interval, the proposed method provides 60\% (SNR=5) and 68\% (SNR=10) accuracy while KPLS provides approximately 68\% for each SNR. It is observed that the high classification accuracy of HDMR is achieved utilizing more features, such as 20-30 features, in classification in the cases when SNR is high (between 5-20). This case is not valid after SNR equals 20, meaning that HDMR achieves the highest performance with relatively fewer features, around 13-14, and KOPLS outperforms the KPLS method. Although KOPLS achieves very good overall accuracy, its AUC is relatively very low in comparison to our proposed method as observed in Fig. \ref{fig:snrKNN}(a),\ref{fig:snrKNN}(c),\ref{fig:snrSVM}(e). This is due to the fact that the learning curves of 1-nn and SVM obtained with KOPLS increase very slowly and provide very low accuracy with the first extracted features in the latent space.

\subsection{Computational Complexity}
The proposed method consists of three steps affecting the computational cost: calculating the graph affinity matrix, the graph Laplacian matrix, and then solving a generalized eigenvalue problem. The computational complexity of the first step requires ${\mathcal O}(pnm^2)$ operations, where $p$, $n$, and $m$ refer to the order of polynomials, the dimension of the ambient space, and the number of training samples. {\color{black} For computing the graph Laplacian matrix, ${\mathcal O}(km^{2})$ operations are needed to find k-nearest neighbors of all the training samples. The complexity of solving the generalized eigenvalue problem is ${\mathcal O}(p^3 n^3)$. To project the samples to the $d$-dimensional latent space, $d$ smallest eigenvalues are computed with the complexity of ${\mathcal O}(d p^{2}n^{2})$. Thus, the total complexity of generalized eigenvalue problem is ${\mathcal O}(p^3 n^3 + dp^{2}n^{2})$.}  Therefore, the algorithm scales as ${\mathcal O}(pnm^2 + km^{2}p^{2} + p^3 n^3 + dp^2n^{2})$, and since typically $m\gg n \gg d \gg k$, the first two terms dominate the overall cost which scales as $\mathcal {O}(nm^2)$. {\color{black} We observed that a sensible order of the Legendre polynomial $p$ is 2 or 3 in most of the applications. Thus it does not affect the total complexity of the proposed method.} {\color{black} The time complexity of all the methods employed in this study is given in table \ref{table:cc}.}

\begin{table}[!h]
\caption{ \color{black} Computational complexities of the manifold learning methods.}
\begin{tabular}{llllll}
\hline
 HDMR & LPP & LSDA & KPLS & KOPLS & LLE \\ \hline
 $\mathcal {O}(nm^2)$   & $\mathcal {O}(nm^2)$  & $\mathcal {O}(n^{3})$     &  $\mathcal {O}(m d^{2})$    &  $\mathcal {O}(m^{3})$     &   $\mathcal {O}(n^{3})$\\
 \hline
\end{tabular}
\label{table:cc}
\end{table}






\section{Conclusions}
\label{sec:foot}
In this paper, we proposed a manifold learning method based on HDMR which \textcolor{black}{extracts  discriminative features from the image and} has the ability to solve the out-of-sample problem, unlike most manifold learning methods. The coordinates of the samples in the latent space are represented by the first-order terms of the HDMR, and the graph embedding framework is modified accordingly. Even though our method holds some similarities with classical LPP, it is different in terms of its nonlinearity assumption, which is ensured by using higher-order Legendre polynomials. This provides the main advantage: the feature space is orthogonally expanded to high dimensional spaces with respect to the order of the LPs so richer feature representations are achieved. To prevent the subspace learning method from overfitting when increasing the order of Legendre polynomials, regularization of the generalized eigenvalue problem turned to be strictly necessary. Since we consider particularly the classification problem in this study, we used the class information in the construction of the graph affinity matrix; hence the proposed method leads to a {\em supervised} manifold learning method. However, it should be noted that the proposed method can be extended to {\em unsupervised} and {\em(semi)supervised} settings as well with a special design of the graph affinity matrix. To demonstrate the effectiveness of the proposed method, we used three hyperspectral \textcolor{black}{datasets} 
 and compared the performance 
to its counterparts \textcolor{black}{favourably}. The experimental results showed that HDMR-based manifold learning provides more representative features in the latent space than 
the other methods \textcolor{black}{and can constitute a defacto choice for feature extraction for remote sensing data analysis}. {\color{black} An interesting possibility to extend the work will consider incorporating spatial information, which can be done in different ways: in defining the graph via kernel composition \cite{marconcini2009composite} or multiscaling \cite{tuia2009semisupervised}, working with graphlets \cite{camps2010spatio} or hypergraphs \cite{camps2009biophysical} and/or correspondingly in defining the embedding.}





\bibliographystyle{IEEEtran}
\bibliography{tgars2020}

\begin{IEEEbiography}[{
\includegraphics[width=1in,height=1.25in,clip,keepaspectratio]{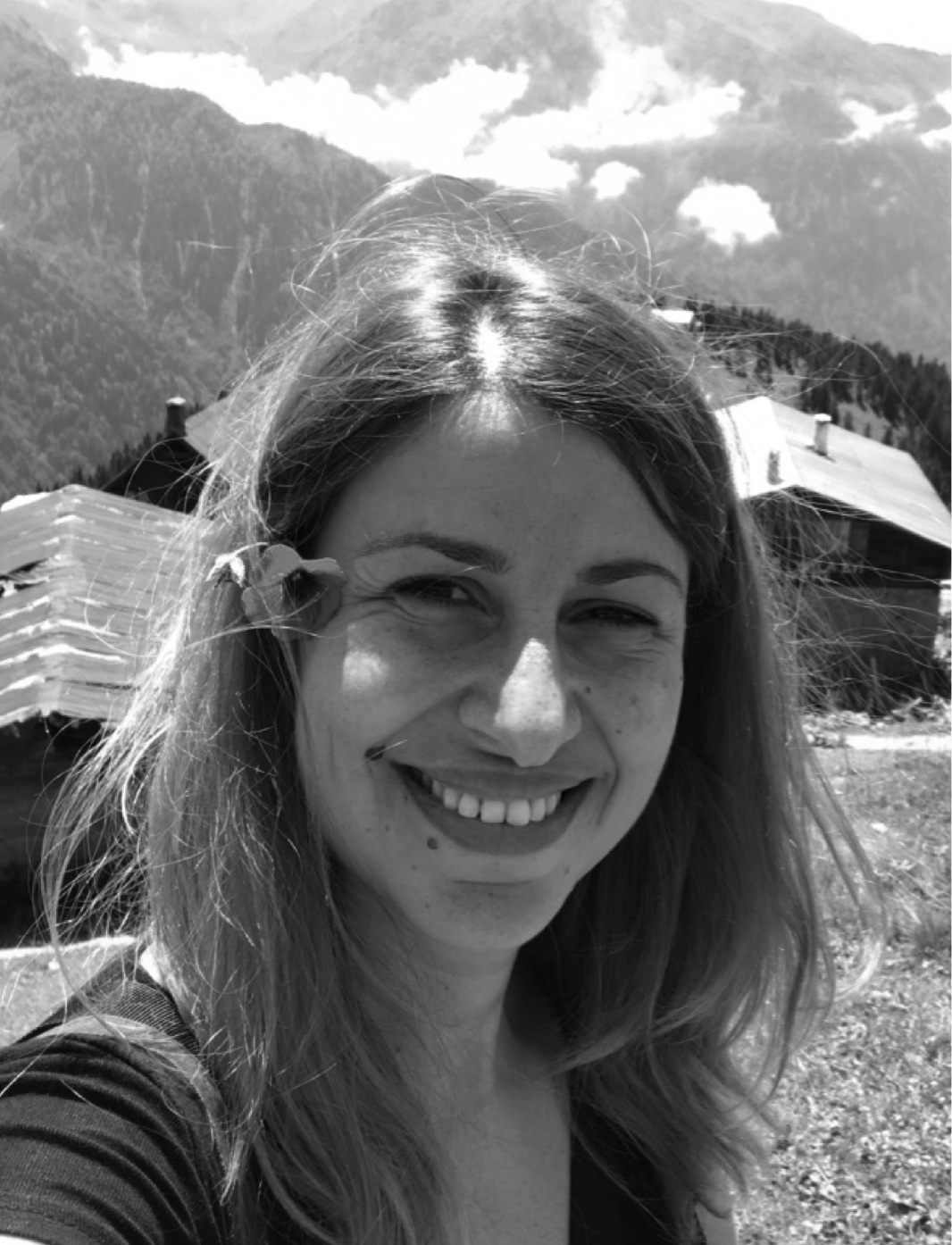}

}]{G\"ul\c sen Ta\c sk\i n}
(S'09-M'10) received the B.S. degree in Geomatics Engineering, the M.S. degree in Computational Science and Engineering (CSE), and the Ph.D. degree in CSE, from Istanbul Technical University (ITU), Turkey, in 2001, 2003, and 2011, respectively.
She is currently an Associate Professor in  Institute of Disaster Management in Istanbul Technical University. She was a visiting scholar at School of Electrical and Computer Engineering and School of Civil Engineering in Purdue University, in 2008-2009 and 2016-2017, respectively. 
 She is a reviewer for Photogrammetric Engineering and Remote Sensing, IEEE Transactions on Geoscience and Remote Sensing,  IEEE Journal of Selected Topics in Applied Earth Observations and Remote Sensing, IEEE Geoscience and Remote Sensing Letters, and IEEE Transactions on Image Processing. Her current interests include machine learning approaches in hyperspectral image analysis, dimensionality reduction, and sensitivity analysis. 
\end{IEEEbiography}
\vspace{-0.3in}
\raggedbottom

\begin{IEEEbiography}[{
\includegraphics[width=1in,height=1.25in,clip,keepaspectratio]{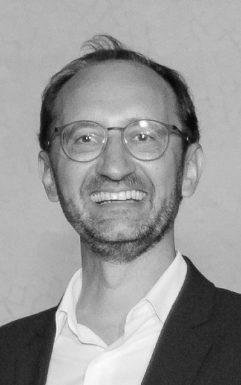}
}]{Gustau Camps-Valls} (F'18) received the Ph.D. degree in physics from the Universitat de València, Val\`encia, Spain, in 2002. He is currently a Full Professor of Electrical Engineering and Coordinator of the Image and Signal Processing (ISP) group, Universitat de Val\`encia. He is involved in the development of machine learning algorithms for geoscience and remote sensing data analysis. He has authored 300 journal papers, more than 250 conference papers, and 25 international book chapters. He holds a Hirsch's index, h = 72 (source: Google Scholar), entered the ISI list of Highly Cited Researchers, in 2011, Thomson Reuters ScienceWatch identified one of his papers on kernel-based analysis of hyperspectral images as a fast moving front research, 10 papers as trending papers, and a paper on deep learning and process understanding as a hot paper in 2019. He is the Editor for the books entitled `Kernel Methods in Bioengineering, Signal and Image Processing' (Hershey, PA, USA: IGI, 2007), `Kernel Methods for Remote Sensing Data Analysis' (Hoboken, NJ, USA: Wiley, 2009), `Remote Sensing Image Processing' (San Rafael, CA, USA: Morgan \& Claypool, 2011), `Digital Signal Processing with Kernel Methods' (Hoboken, NJ, USA: Wiley, 2018), and `Deep Learning for the Earth Sciences'  (Hoboken, NJ, USA: Wiley, 2021). Prof. Camps-Valls is a recipient of two prestigious European Research Council (ERC) grants; a consolidator grant on Statistical Learning for Earth Observation Data Analysis, in 2015, and a Synergy grant on `Understanding and Modeling the Earth System with Machine Learning', in 2019. He is/has been an Associate Editor for the IEEE Transactions on Signal Processing, IEEE Geoscience and Remote Sensing Letters, and IEEE Signal Processing Letters. He was the Invited Guest Editor for the IEEE Journal of Selected Topics in Signal Processing, in 2012 and IEEE Geoscience and Remote Sensing Magazine, in 2015.  In 2016 he was included in the prestigious IEEE Distinguished Lecturer program of the GRSS.
\end{IEEEbiography}

\end{document}